%% file: main.tex
\documentclass[11pt]{article}

\usepackage[preprint]{acl}
\usepackage{times}
\usepackage{latexsym}
\usepackage[T1]{fontenc}
\usepackage[utf8]{inputenc}
\usepackage{microtype}
\usepackage{inconsolata}
\usepackage{graphicx}
\usepackage{amsmath}
\usepackage{amssymb}
\usepackage{booktabs}
\usepackage{placeins}
\usepackage{dblfloatfix}
\title{SnakeTranslate: Weakly Supervised Policy Learning from Offline Parallel Text for Simultaneous Translation}
\author{
 \textbf{Yu Pu\textsuperscript{$\dagger$}},
 \textbf{Zhiyun Fan},
 \textbf{Baiji Liu},
 \textbf{Zhifu Gao},
 \textbf{Xiangang Li}
 \\
 \\
 Alibaba Token Hub, Alibaba Group\\
 \small{\textsuperscript{$\dagger$}Corresponding author.}\\
 \small{\textbf{Correspondence:} \href{mailto:david.py@alibaba-inc.com}{david.py@alibaba-inc.com}}
}

\begin{document}
\maketitle

\input{sections/abstract}
\input{sections/introduction}
\input{sections/related_work}
\input{sections/method}
\input{sections/experimental_setup}
\input{sections/results}
\input{sections/conclusion}
\input{sections/limitations}

\FloatBarrier

\input{references.bbl}
\clearpage
\appendix
\input{sections/result_coverage}
\input{sections/appendix_cases}

\input{sections/appendix_architecture}
\input{sections/appendix_alignment}
\input{sections/implementation_details}
\input{sections/appendix_evaluation}
\FloatBarrier

\end{document}

%% file: sections/abstract.tex
\begin{abstract}
Adaptive simultaneous translation requires reliable prefix translation and timely READ/WRITE decisions, but ordinary parallel corpora provide no action annotations. We present SnakeTranslate, which adapts loss-derived alignment to sampled causal histories for jointly learning both capabilities. The framework supports supervised fine-tuning (SFT) on parallel text and on-policy distillation (OPD) with reverse Kullback--Leibler (KL) divergence on student-generated translations. At inference, source and target tokens form one append-only interleaved history, enabling incremental decoding with cache reuse. On FLEURS, NTREX128, and WMT24++, our 4B model improves quality--latency trade-offs over streaming policies applied to the same offline-tuned backbone, with further gains from OPD. It also compares favorably with larger translation models at low and moderate latency, while higher-latency quality gaps remain on WMT24++.
\end{abstract}

%% file: sections/introduction.tex
\section{Introduction}
\label{sec:introduction}

Simultaneous machine translation (SimulMT) alternates between reading source tokens and writing target tokens \citep{dalvi2018incremental}. More context resolves ambiguity and reordering but increases latency; early output risks incorrect commitments. Training on mismatched prefixes can encourage predictions before their supporting source content arrives \citep{chen2021pseudoreferences,ma2023nast}. Reliable prefix translation and adaptive READ/WRITE policies are therefore complementary requirements.

\begin{figure*}[tp]
 \centering
 \includegraphics[width=\textwidth]{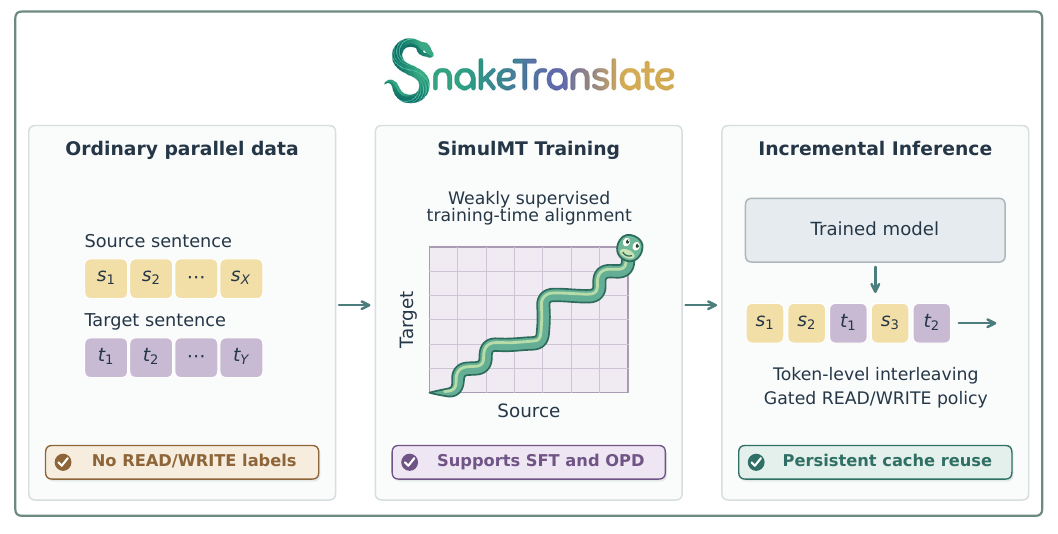}
 \caption{SnakeTranslate learns from parallel sentences without READ/WRITE annotations. Supervised fine-tuning (SFT) and on-policy distillation (OPD) share loss-derived alignment supervision. The trained model follows one append-only history with cache reuse.}
 \label{fig:intro_overview}
\end{figure*}

Ordinary parallel corpora lack READ/WRITE annotations. Existing methods use fixed schedules \citep{ma2019stacl}, construct aligned trajectories \citep{koshkin2024transllama,fu2025east}, or learn adaptive policies from sentence-level objectives \citep{liu2021caat,zhang2022dualpaths,guo2024dst}. Attention patterns and prediction stability also supply timing signals \citep{papi2023edatt,zhao2025drfrattn,zhao2024psfuture}. We build on these insights to derive a structured boundary from prediction costs, jointly determining when to write and which prefix states receive translation supervision.

We present \textbf{SnakeTranslate}, which adapts loss-derived alignment to a causal language-model backbone and extends it to distillation. Two-dimensional source--target modeling has earlier precedents \citep{bahar2018twodimensional}. \citet{pu2026dualstream} introduced loss-heatmap dynamic programming (DP), dense policy labels, and region-restricted translation loss for a dual-stream grid. SnakeTranslate transfers this supervision to sampled causal histories, enabling execution along one append-only history. Reference-token costs determine the supervised boundary; distillation extends the same principle to student-generated target prefixes. Table~\ref{tab:prior_comparison} summarizes the architectural and training differences.

In decoder-only large language models (LLMs), inserting new source text before an existing target prefix invalidates cached context \citep{raffel2024simulmask}. Following cache-compatible streaming designs \citep{guo2024dst,wang2025conversational,fu2025east}, SnakeTranslate instead appends source and target tokens to one interleaved history. Each step updates recurrent states and attention key--value (KV) caches, with computation linear in history length for fixed model dimensions.

After supervised fine-tuning (SFT), on-policy distillation (OPD) uses a frozen full-source teacher to score student continuations \citep{agarwal2024gkd}. Reverse Kullback--Leibler (KL) divergence replaces reference-token loss in both boundary inference and translation learning (Figure~\ref{fig:intro_overview}).

On FLEURS, NTREX128, and WMT24++, our 4B model improves over policies applied to the same offline-tuned initialization, with further gains from OPD. SFT and OPD introduce no new source data beyond offline training. Comparisons with external systems show favorable low- and moderate-latency performance, with remaining high-latency quality gaps on WMT24++.

Our contributions are threefold:
\begin{itemize}
 \item A sampled-history causal lattice that adapts dual-stream loss-derived alignment to a pretrained causal language model with single-history incremental inference.
 \item Reverse-KL OPD coupling boundary inference and translation learning on student-generated target prefixes.
 \item Evaluation on three public benchmarks against shared-initialization policies and external systems over matched directions.
\end{itemize}

%% file: sections/related_work.tex
\section{Related Work}
\label{sec:related_work}

Prefix alignment supplies segmentation supervision for SimulMT \citep{kano2022prefix}.

Wait-$k$ trains prefix translation under fixed schedules \citep{ma2019stacl}; multi-path training shares a model across latency settings \citep{elbayad2020efficient}. LLM-based methods construct interleaved trajectories: TransLLaMa derives action labels from causal alignment \citep{koshkin2024transllama}, and Efficient and Adaptive Simultaneous machine Translation (EAST) builds latency-aware chunks \citep{fu2025east}. Conversational SimulMT similarly uses aligned source--target chunks for cache reuse \citep{wang2025conversational}. SimulMask supports cache-compatible prefix translation through policy-conditioned masks and positional handling, with wait-$k$ experiments; adaptive policy learning remains separate \citep{raffel2024simulmask}.

Latent or differentiable decisions offer another route. Monotonic Multihead Attention and information-transport policies learn source access without annotated actions \citep{ma2020mma,zhang2022itst}. The Cross Attention Augmented Transducer (CAAT) marginalizes READ/WRITE paths, while the Hidden Markov Transformer (HMT) models latent WRITE positions \citep{liu2021caat,zhang2023hmt}. Non-autoregressive streaming models also combine latent alignment and latency losses \citep{ma2023nast}. The Decoder-only Streaming Transformer (DST) learns source sufficiency using separate stream positions and Streaming Self-Attention, without re-encoding target prefixes \citep{guo2024dst}.

Predictions can also supply policy supervision. Reference-token ranks yield greedy oracle actions \citep{zheng2019simpler}, and candidate translations guide adaptive segmentation \citep{zhang2020segmentation}. Divergence-based adaptive policy (DaP) compares full-input and prefix predictions \citep{zhao2023dap}; Self-Modifying State Modeling (SM$^2$) learns confidence through their mixture \citep{yu2024sm2}. In speech translation, Regularized Entropy INformation Adaptation (REINA) uses reference-loss differences \citep{hirschkind2026reina}, while SimulMEGA learns policy routing and selects confident positions for prefix loss \citep{le2025simulmega}.

Local Agreement provides a training-free policy, committing stable prefixes across hypotheses from successive input updates \citep{liu2020localagreement}.

SnakeTranslate adapts the cost-derived boundary supervision of \citet{pu2026dualstream} to sampled causal histories and extends it to student-generated target prefixes with reverse-KL distillation (Section~\ref{sec:opd}).

%% file: sections/method.tex
\section{SnakeTranslate}
\label{sec:method}

\subsection{Model and Incremental Inference}
\label{sec:model_architecture}
\label{sec:inference}

The source $s_{1:X}$ and translation $t_{1:Y}$ include end-of-sequence (EOS) markers. State $(x,y)$ consumes $s_{1:x}$ and $t_{1:y}$ and predicts $t_{y+1}$; training covers $0\leq x\leq X$ and $0\leq y<Y$. Task-prompt tokens provide context but are omitted from this notation and excluded from supervision.

\begin{figure*}[tp]
 \centering
 \includegraphics[width=\textwidth]{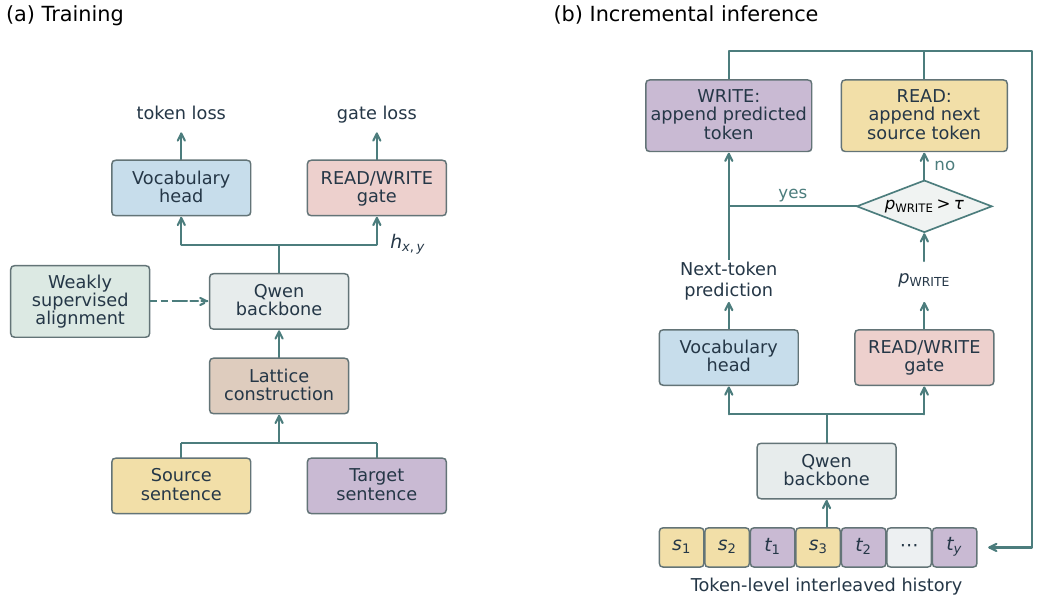}
 \caption{Shared training and inference architecture. (a) Unaligned parallel sentences form a lattice; the dashed arrow denotes weak alignment supervision. (b) Inference appends a predicted target token when WRITE probability exceeds $\tau$, or the next source token otherwise, updating the persistent cache.}
 \label{fig:model_architecture}
\end{figure*}

The Qwen backbone \citep{qwen2026qwen35} feeds each final normalized hidden vector $h_{x,y}$ to a vocabulary head and a two-action multilayer perceptron (MLP) gate (Figure~\ref{fig:model_architecture}; Appendix~\ref{app:architecture}).

Training constructs a causal lattice from unaligned sentence pairs, passing one hidden vector per cell between layers. Adapting loss-derived alignment to this causal lattice and to distillation is our central contribution; Section~\ref{sec:weak_alignment} describes the inherited DP supervision, and Section~\ref{sec:opd} gives its KL extension.

At inference, WRITE probability above $\tau$ appends a predicted target token; otherwise, READ appends the next source token. After source completion, generation continues until EOS or the output limit. Both heads share one cached interleaved history, requiring $O(L)$ attention work per step for history length $L$ and fixed model dimensions (Appendix~\ref{app:execution_cost}). Adjusting $\tau$ controls latency without retraining.

\subsection{Lattice Representations}
\label{sec:prefix_states}
\label{sec:grid_execution}

\begin{figure*}[tp]
 \centering
 \includegraphics[width=0.9\textwidth]{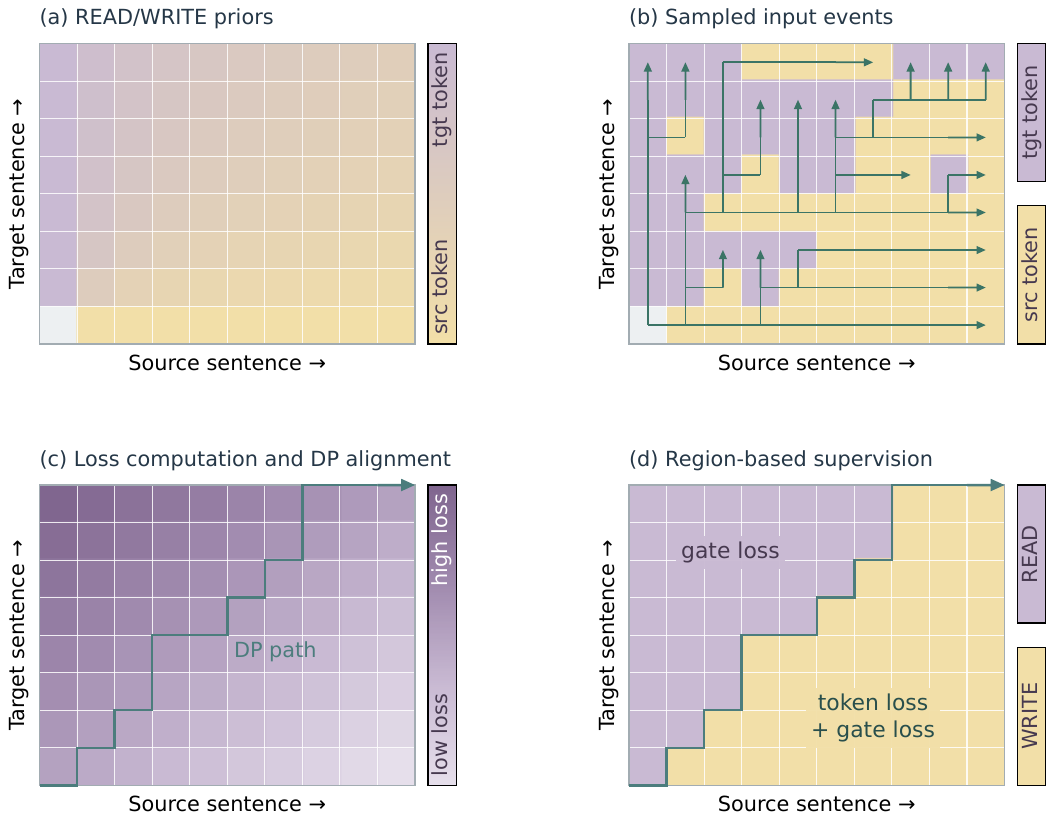}
 \caption{Lattice training: (a) READ/WRITE priors; (b) a sampled causal history tree; (c) prediction costs and DP alignment; (d) gate supervision throughout the grid and token supervision below the boundary. Source and target prefixes increase rightward and upward. Yellow/purple denote source/target tokens; gray marks the root. Green edges share prefixes. The heatmap and alignment are schematic.}
 \label{fig:training_overview}
\end{figure*}

Figure~\ref{fig:training_overview} summarizes lattice training.

Each non-root cell $(x,y)$ samples a predecessor shared across layers: READ appends $s_x$ to $(x-1,y)$ with probability $x/(x+y)$; WRITE appends $t_y$ to $(x,y-1)$ with probability $y/(x+y)$. Axis cells use their valid predecessor. Tracing back to the shared root defines history $H_{x,y}$, preserving within-stream order and sharing prefix computations (Figure~\ref{fig:training_overview}(a--b)).

The initial vector is $E(s_x)$ or $E(t_y)$. Rotary position embeddings \citep{su2021roformer} use within-stream indices in training and inference; role embeddings are disabled.

Gated DeltaNet \citep{yang2025gated} applies convolution and recurrence along sampled parents. Gated grouped-query attention \citep{ainslie2023gqa} accesses only the ancestor path, including the current cell. Layer-local recurrent states and attention caches cannot access siblings or unread source tokens.

For a fixed path, each layer uses the same ordered tokens, stream-local positions, and ancestors as one-dimensional causal execution. This is an operation-level correspondence with stochastic layers disabled, not invariance across different interleavings. The sampled computation tree need not coincide with the DP supervision path below.

\subsection{Weakly Supervised Alignment}
\label{sec:weak_alignment}
\label{sec:dp_supervision}
\label{sec:joint_learning}

Let $p_{x,y}$ and $a_{x,y}$ denote the vocabulary distribution and WRITE probability obtained from $h_{x,y}=f_\theta(H_{x,y})$.
During SFT, the cost is the reference-token cross-entropy (CE):
\begin{equation}
 c^{\mathrm{SFT}}_{x,y}=-\log p_{x,y}(t_{y+1}).
 \label{eq:sft_cost}
\end{equation}
The heatmap $\mathbf C\in\mathbb R^{(X+1)\times Y}$ compares target-token costs across source prefixes. A detached copy of all candidate costs determines a monotone supervision boundary.

For a monotone READ/WRITE path $\pi$ from $(0,0)$ to $(X,Y)$, let $r_j$ be the source-prefix length when writing $t_j$, and $g_x$ the target count before leaving source prefix $x$, with $g_X=Y$. Its cost and completed-output area are
\begin{align}
 C(\pi)&=\sum_{j=1}^{Y}c_{r_j,j-1}, \\
 A(\pi)&=\sum_{x=0}^{X}g_x
       =\sum_{j=1}^{Y}(X+1-r_j).
 \label{eq:area}
\end{align}
We reward earlier output by solving
\begin{equation}
 \pi^\star=\arg\max_{\pi}\{\lambda A(\pi)-C(\pi)\},
 \qquad \lambda>0.
 \label{eq:path_objective}
\end{equation}
Reading more context can lower prediction cost but sacrifices area reward.

DP solves this in $O(XY)$ time (Figure~\ref{fig:training_overview}(c)). Excluding the constant terminal reward, the best score is
\begin{equation}
 F(x,y)=\max\!\left\{
 \begin{aligned}
  &F(x-1,y)+\lambda y,\\
  &F(x,y-1)-c_{x,y-1}.
 \end{aligned}\right\}
 \label{eq:dp}
\end{equation}
The entries correspond to READ and WRITE. Backtracking yields $g_x^\star$ and the mask $M_{x,y}=\mathbf{1}[y<g_x^\star]$, labeling cells below the boundary WRITE and the rest READ (Figure~\ref{fig:training_overview}(d)).

DP exactly optimizes this fixed cost map for a given $\lambda$; it does not optimize over the histories used to compute the costs. Appendices~\ref{app:alignment_details} and~\ref{sec:alignment_case} give boundary conventions and a reordering example.

For valid cells $\mathcal V$, excluding padding and target prompts, binary cross-entropy (BCE) supervises the gate throughout the grid, while translation uses only the WRITE region:
\begin{align}
 \mathcal L_{\mathrm{gate}}&=\frac{1}{|\mathcal V|}
 \sum_{(x,y)\in\mathcal V}\operatorname{BCE}(a_{x,y},M_{x,y}),
 \label{eq:gate_loss}
 \\
 \mathcal L_{\mathrm{trans}}&=
 \frac{\sum_{(x,y)\in\mathcal V}M_{x,y}c_{x,y}}
      {\sum_{(x,y)\in\mathcal V}M_{x,y}},
 \label{eq:translation_loss}
 \\
 \mathcal L_{\mathrm{SFT}}&=\mathcal L_{\mathrm{trans}}
                         +\mathcal L_{\mathrm{gate}}.
\end{align}
The mask is fixed during backpropagation and regenerated during training. READ cells receive only policy and shared-computation gradients, reducing pressure to translate from insufficient input. SFT pools valid cells across a minibatch; OPD instead averages per-example losses (Appendix~\ref{app:opd_reduction}). The boundary remains a weak signal of source sufficiency.

\subsection{On-Policy Distillation}
\label{sec:opd}

In OPD \citep{agarwal2024gkd}, the SFT-initialized student generates $\hat t\sim\operatorname{Rollout}_{\theta_k}(s)$. A frozen teacher scores this continuation with the complete source in a single causal pass:
\begin{equation}
 q_y(v)=p_T(v\mid s_{1:X},\hat t_{1:y}),
 \quad 0\leq y<|\hat t|.
 \label{eq:teacher}
\end{equation}
Each $q_y$ is shared across source-prefix states.

We construct a newly sampled history tree from $s$ and $\hat t$, replacing Eq.~\ref{eq:sft_cost} with reverse Kullback--Leibler (KL) divergence, also used in language-model distillation \citep{gu2024minillm}:
\begin{equation}
 \begin{split}
 c^{\mathrm{OPD}}_{x,y}
 &=D_{\mathrm{KL}}(p_{x,y}\Vert q_y)\\
 &=\sum_{v\in\mathcal U}p_{x,y}(v)
       \big[\log p_{x,y}(v)-\log q_y(v)\big],
 \end{split}
 \label{eq:reverse_kl}
\end{equation}
Here $\mathcal U$ is the full vocabulary, compared by token ID. Base IDs agree; Appendix~\ref{app:teacher_scoring} discusses tokenizer and special-token differences.

Detached KL costs induce the boundary through Eq.~\ref{eq:path_objective} and train translation within its WRITE region. OPD averages the unit-weighted translation and gate losses per example, without reference-token loss (Appendix~\ref{app:opd_reduction}).

Rollouts use updated student weights; their target tokens, not their READ/WRITE histories, supply the training examples. Rollout actions, teacher scoring, and DP backtracking receive no gradients. DP labels are recovered from the new cost map, not copied from rollout actions. Teacher information enters only through supervision; student states retain causal prefixes. Table~\ref{tab:training_procedure} summarizes both stages.

%% file: sections/experimental_setup.tex
\section{Experimental Setup}
\label{sec:experimental_setup}

\subsection{Training}
\label{sec:training_setup}

We use approximately 171.6 million internal multilingual parallel records, including duplicates, with language labels but no READ/WRITE annotations. Each training stream is limited to 128 positions; Appendix~\ref{app:task_formatting} gives the prompt and EOS format.

Qwen3.5-4B-Base is first fine-tuned for offline translation for 5,000 updates. This checkpoint initializes SnakeTranslate and the Qwen policy baselines. Subsequent SFT and OPD use data already covered by offline training, introducing no new source data.

SFT jointly trains the backbone and gate with AdamW \citep{loshchilov2019adamw}, learning rate $10^{-5}$, and effective batch size 4,096. We evaluate after 540 updates and initialize OPD from this checkpoint.

OPD uses a frozen internal Qwen3.8 teacher with reasoning disabled, constant AdamW learning rate $10^{-6}$, and effective batch size 1,024. Rollouts use greedy vLLM decoding \citep{kwon2023vllm} at $\tau=0.95$. We evaluate after 2,000 updates, totaling 2.048 million sample updates.

Only OPD uses the teacher. The same SFT and OPD checkpoints are evaluated across all benchmarks and directions. \textit{Ours} denotes OPD except in stage comparisons. Appendix~\ref{app:dp_details} gives numerical and scoring details.

\subsection{Evaluation}
\label{sec:evaluation_setup}

We evaluate on FLEURS transcriptions \citep{conneau2022fleurs}, NTREX-128 \citep{federmann2022ntrex} (NTREX128 below), and WMT24++ \citep{deutsch2025wmt24pp}. Shared-backbone comparisons cover 42/42/7 directions, while offline and external-model comparisons use their common 13/13/7 directions, with identical instances within each direction (Appendix~\ref{app:evaluation_details}).
We apply wait-$k$, Relaxed Agreement Longest Common Prefix (RALCP; \citealp{wang2024ralcp}), and Local Agreement (LA; \citealp{liu2020localagreement}) to offline-tuned Qwen3.5-4B, Hy-MT2-30B-A3B \citep{zheng2026hymt2}, TranslateGemma-27B-it \citep{finkelstein2026translategemma}, and Riva-Translate-4B-Instruct-v2 \citep{nvidia2026riva}. Wait-$k$/RALCP use $k=3,6,9$; LA uses read increments $c=1,2,3$. Specialized systems include NoLanguageLeftWaiting (NLLW),\footnote{\url{https://github.com/QuentinFuxa/NoLanguageLeftWaiting}} toLLMatch-8B/70B \citep{koshkin2024tollmatch}, Simul-LLM \citep{agostinelli2024simulllm}, and EAST-8B \citep{fu2025east}. Their native policies and coverage are detailed in Appendix~\ref{app:baseline_protocols}. All systems treat EOS before source completion as READ.
Offline controls evaluate Qwen3.5-4B-Base, its offline-tuned checkpoint, Riva, TranslateGemma, and Hy-MT2 with complete sources on the same 13/13/7 intersection.

We report BLEURT-20 \citep{sellam2020bleurt,pu2021compactmetrics} and XCOMET-XL \citep{guerreiro2024xcomet}, labeled BLEURT and COMET, and length-adaptive average lagging (LAAL; \citealp{papi2022laal}) in common FLoRes subword units. All metrics are averaged within directions and then equally across directions. Empty and capped outputs retain quality scores; empty outputs have undefined latency. LAAL excludes computation time (Appendix~\ref{app:latency_protocol}).

SnakeTranslate uses persistent-cache vLLM inference with greedy decoding, batch size 128, and $\tau=0.8,0.9,0.95$.

%% file: sections/results.tex
\section{Results and Analysis}
\label{sec:results}

\subsection{Offline Translation}
\label{sec:offline_results}

Table~\ref{tab:offline_results} compares the base and offline-tuned Qwen models with external translation systems using complete sources and matched directions. It measures the benefit of offline adaptation and the initialization's translation capability.

\begin{table}[!tp]
 \centering
 \fontsize{8}{10}\selectfont
 \setlength{\tabcolsep}{1.5pt}
 \begin{tabular*}{\columnwidth}{@{\extracolsep{\fill}}lccc@{}}
 \toprule
 Model & FLEURS & NTREX128 & WMT24++\\
 \midrule
 \input{tables/offline_common_rows}
 \end{tabular*}
 \caption{Offline BLEURT/COMET on 13/13/7 common directions for FLEURS/NTREX128/WMT24++. Ours (offline) is the offline-tuned Qwen3.5-4B-Base. The divider separates 4B from larger models; Hy-MT2 has 30B total and 3B active parameters. Directions receive equal weight.}
 \label{tab:offline_results}
\end{table}

Offline adaptation improves BLEURT by 0.025--0.039 and COMET by 0.031--0.059 across the three benchmarks, surpassing Riva-4B on both metrics. TranslateGemma and Hy-MT2 remain stronger, particularly on WMT24++ (Appendix~\ref{app:benchmark_analysis}). These gaps establish the 4B initialization's capability relative to larger backbones before streaming constraints are imposed.

\input{sections/training_stage_ablation}

\subsection{External Translation Systems}
\label{sec:external_results}

Figure~\ref{fig:external_systems} compares Ours with wait-$k$, RALCP, and LA-2 on external backbones over their common 13/13/7 directions.

On FLEURS, Ours reaches BLEURT/COMET of 0.757/0.935 at LAAL 6.676, exceeding TranslateGemma with RALCP (0.730/0.902 at 6.638). On NTREX128, it improves both metrics while reducing LAAL from 8.516 to 7.261.

On WMT24++, Ours exceeds TranslateGemma with RALCP at low latency. At higher delay, TranslateGemma has higher COMET, and Hy-MT2 exceeds Ours on both metrics, consistent with the offline backbone quality gaps.

LA-2 is stronger: Hy-MT2 with $c=3$ on FLEURS trades higher COMET for greater delay. TranslateGemma with $c=2$ on WMT24++ reaches 0.661/0.828 at LAAL 5.698, exceeding Ours at $\tau=0.9$ (0.650/0.792 at 5.513). Appendix~\ref{app:result_coverage} analyzes benchmark composition and individual directions.

\subsection{Specialized Simultaneous Systems}
\label{sec:direction_results}

Table~\ref{tab:specialized_mean} compares specialized systems over their differing direction coverage. Within each direction, systems use identical instances and benchmark sets, averaged with equal benchmark weights. Comparisons use measured LAAL because native policy parameters have different meanings.

Ours improves both quality metrics and reduces LAAL against NLLW in all four shared directions at every reported threshold, and against all reported Simul-LLM settings on English-to-Spanish. toLLMatch-70B is stronger, but Ours at $\tau=0.8$ still reaches 0.710/0.875 at LAAL 3.100 on English-to-Spanish, compared with 0.698/0.870 at 7.807 for its $k=3$ setting. Lowering toLLMatch's $k$ reduces delay at a cost in quality.

EAST learns its adaptive policy from explicitly segmented data. On English-to-Chinese, Ours at $\tau=0.8$ reaches 0.681/0.813 at LAAL 5.478, compared with 0.647/0.751 at 10.278 for EAST's \textit{medium-high} prompt; the same threshold also improves both quality metrics with less delay on Chinese-to-English. Appendix~\ref{app:result_coverage} provides per-benchmark results and coverage details.

Appendix~\ref{app:case_studies} complements these quantitative comparisons with a loss-derived alignment example and streaming traces at three thresholds.

%% file: tables/offline_common_rows.tex
Ours (offline) & 0.760/0.940 & 0.727/0.901 & 0.663/0.819 \\
Qwen3.5-4B-Base & 0.735/0.908 & 0.692/0.849 & 0.625/0.760 \\
Riva-4B & 0.748/0.923 & 0.720/0.885 & 0.639/0.780 \\
\midrule
TranslateGemma-27B & 0.767/0.956 & 0.733/0.921 & 0.679/0.857 \\
Hy-MT2-30B-A3B & 0.772/0.955 & 0.740/0.923 & 0.697/0.870 \\
\bottomrule

%% file: sections/training_stage_ablation.tex
\subsection{Training Stages}
\label{sec:shared_backbone_results}

Figure~\ref{fig:shared_backbone} compares policies on the offline-tuned Qwen model with its SFT and OPD continuations over 42/42/7 directions, using shared source-data coverage. LA uses agreement over two successive input updates and beam size 2 (LA-2), with read increments $c=1,2,3$. Lines connect measured operating points; supplementary \texttt{results\_figure\_data.csv} provides their full-precision coordinates for Figures~\ref{fig:shared_backbone} and~\ref{fig:external_systems}.

\input{sections/results_figures}
\input{tables/specialized_mean}

SFT improves both quality metrics and reduces LAAL relative to RALCP at $k=9$ on all three benchmarks when $\tau=0.95$, supporting joint learning of prefix translation and its policy. The quality gains are smaller on WMT24++.

LA-2 is a stronger baseline. On FLEURS, OPD at $\tau=0.9$ reaches BLEURT/COMET of 0.731/0.889 at LAAL 5.832, compared with 0.723/0.878 at 9.049 for LA-2 with $c=3$. OPD exceeds this highest-quality reported LA-2 setting on both FLEURS and NTREX128 while reducing LAAL by approximately 36\% and 32\%, respectively. On WMT24++, it also improves over LA-2 with $c=2$, but LA-2 with $c=3$ attains higher quality at greater delay.

Across all nine dataset--threshold combinations, OPD improves both quality metrics and reduces LAAL relative to SFT at the same threshold. Aggregate improvements coexist with language-specific regressions; Appendix~\ref{app:stage_details} reports the direction-level counts and English-to-Thai exception. This stage comparison does not isolate the effect of each model component or of removing offline adaptation.

%% file: sections/results_figures.tex
\begin{figure*}[p]
 \centering
 \includegraphics[width=0.94\textwidth]{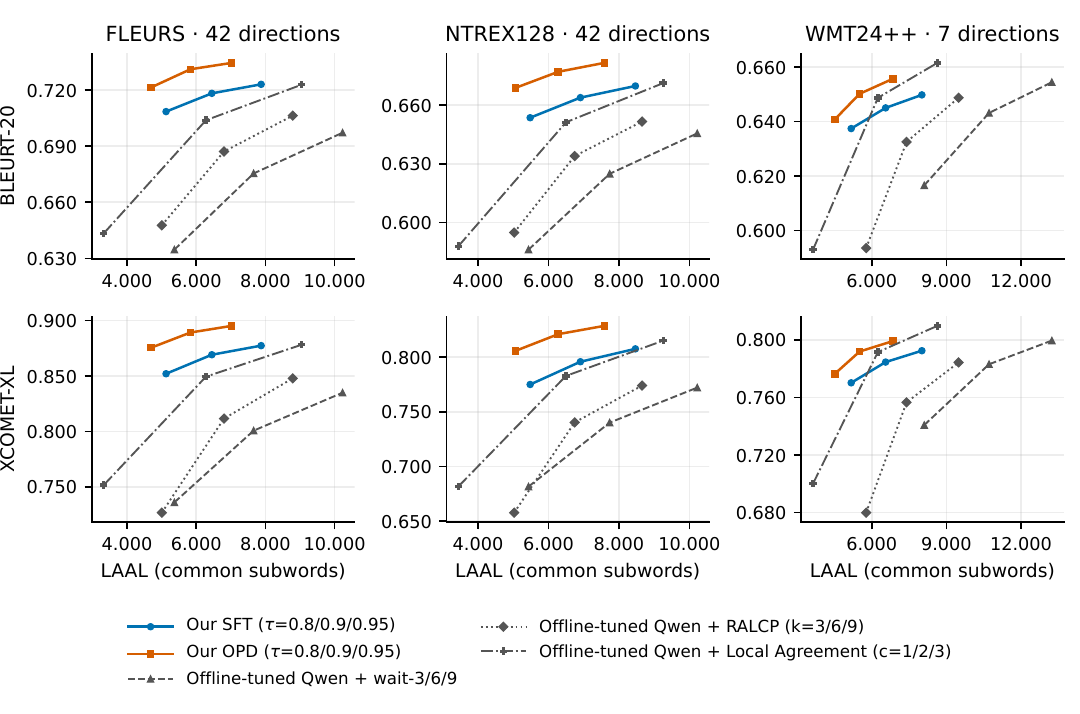}
 \caption{Training-stage comparison. Wait-$k$, RALCP, and LA-2 use the offline-tuned 4B model; SFT starts from that model, and OPD continues from SFT. Each panel averages the same directions and evaluation instances. Lines connect measured operating points.}
 \label{fig:shared_backbone}
 \par\vspace{8pt}
 \includegraphics[width=0.94\textwidth]{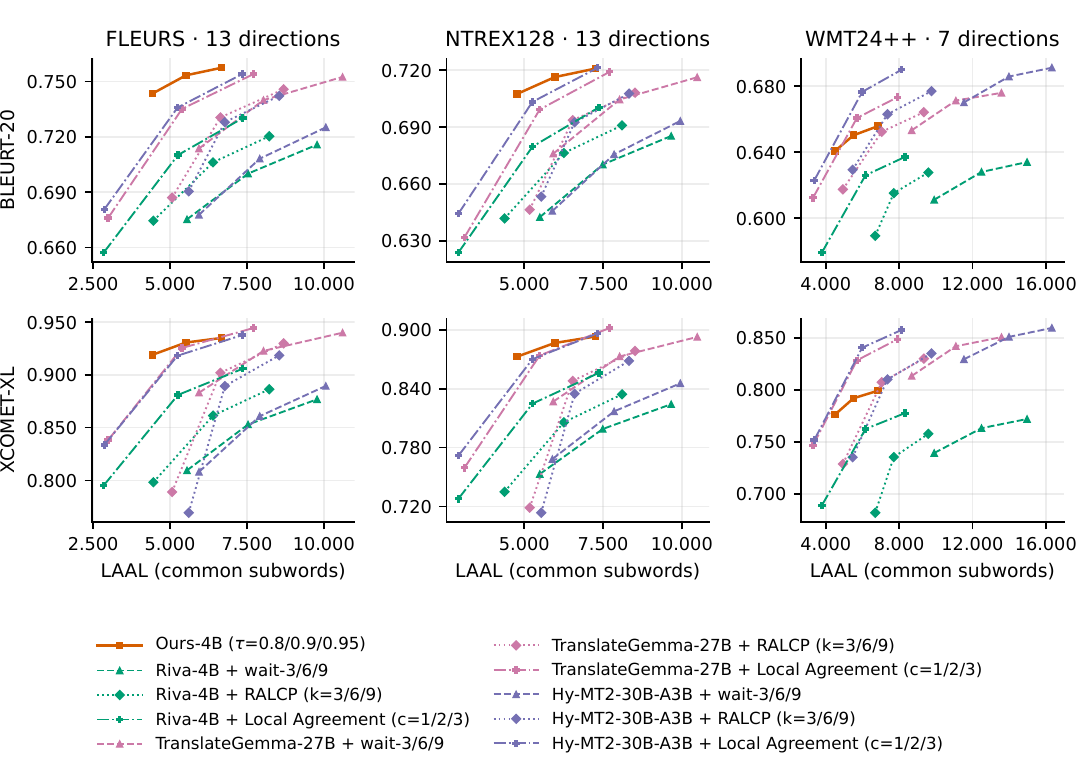}
 \caption{Comparison with external translation systems on 13/13/7 common directions. Ours is compared with Riva-4B, TranslateGemma-27B, and Hy-MT2-30B-A3B under wait-$k$, RALCP, and LA-2. Curves report direction-balanced quality and LAAL in common subword units.}
 \label{fig:external_systems}
\end{figure*}

%% file: tables/specialized_mean.tex
\begin{table*}[tp]
\centering\scriptsize
\setlength{\tabcolsep}{1.5pt}
\resizebox{\textwidth}{!}{%
\begin{tabular}{llccccccccccccccc}
\toprule
System & Setting & \multicolumn{3}{c}{en$\to$es} & \multicolumn{3}{c}{en$\to$ja} & \multicolumn{3}{c}{en$\to$zh} & \multicolumn{3}{c}{es$\to$en} & \multicolumn{3}{c}{zh$\to$en} \\
\cmidrule(lr){3-5}\cmidrule(lr){6-8}\cmidrule(lr){9-11}\cmidrule(lr){12-14}\cmidrule(lr){15-17}
 & & BLEURT & XCOMET & LAAL & BLEURT & XCOMET & LAAL & BLEURT & XCOMET & LAAL & BLEURT & XCOMET & LAAL & BLEURT & XCOMET & LAAL \\
\midrule
Ours & $\tau=0.8$ & 0.710 & 0.875 & \textbf{3.100} & 0.660 & 0.834 & \textbf{6.452} & 0.681 & 0.813 & \textbf{5.478} & 0.756 & 0.941 & \textbf{3.018} & 0.732 & 0.918 & 4.639 \\
 & $\tau=0.9$ & 0.725 & 0.893 & 3.818 & 0.669 & 0.848 & 7.749 & 0.688 & 0.828 & 6.660 & 0.766 & 0.952 & 4.011 & 0.739 & 0.927 & 5.866 \\
 & $\tau=0.95$ & \textbf{0.730} & \textbf{0.899} & 4.621 & \textbf{0.672} & \textbf{0.854} & 9.273 & \textbf{0.694} & \textbf{0.834} & 8.011 & \textbf{0.771} & \textbf{0.955} & 4.951 & \textbf{0.743} & \textbf{0.931} & 7.066 \\
\midrule
toLLMatch-70B & $k=1$ & 0.678 & 0.838 & 5.872 & -- & -- & -- & -- & -- & -- & 0.739 & 0.923 & 3.853 & -- & -- & -- \\
 & $k=2$ & 0.692 & 0.859 & 6.720 & -- & -- & -- & -- & -- & -- & 0.746 & 0.931 & 4.523 & -- & -- & -- \\
 & $k=3$ & 0.698 & 0.870 & 7.807 & -- & -- & -- & -- & -- & -- & 0.754 & 0.939 & 5.437 & -- & -- & -- \\
\midrule
toLLMatch-8B & $k=1$ & 0.630 & 0.772 & 5.889 & -- & -- & -- & -- & -- & -- & 0.687 & 0.834 & 3.675 & -- & -- & -- \\
 & $k=2$ & 0.655 & 0.810 & 6.797 & -- & -- & -- & -- & -- & -- & 0.720 & 0.883 & 4.481 & -- & -- & -- \\
 & $k=3$ & 0.667 & 0.829 & 7.866 & -- & -- & -- & -- & -- & -- & 0.738 & 0.908 & 5.413 & -- & -- & -- \\
\midrule
EAST-8B & low-medium & -- & -- & -- & -- & -- & -- & 0.623 & 0.696 & 5.497 & -- & -- & -- & 0.679 & 0.828 & \textbf{3.314} \\
 & medium & -- & -- & -- & -- & -- & -- & 0.639 & 0.731 & 7.547 & -- & -- & -- & 0.703 & 0.869 & 4.686 \\
 & medium-high & -- & -- & -- & -- & -- & -- & 0.647 & 0.751 & 10.278 & -- & -- & -- & 0.712 & 0.885 & 5.838 \\
\midrule
NLLW & -- & 0.565 & 0.644 & 6.785 & 0.449 & 0.577 & 11.408 & 0.490 & 0.610 & 10.081 & -- & -- & -- & 0.659 & 0.795 & 19.626 \\
\midrule
Simul-LLM & $k=1$ & 0.488 & 0.549 & 4.845 & -- & -- & -- & -- & -- & -- & -- & -- & -- & -- & -- & -- \\
 & $k=2$ & 0.530 & 0.613 & 5.767 & -- & -- & -- & -- & -- & -- & -- & -- & -- & -- & -- & -- \\
 & $k=3$ & 0.550 & 0.646 & 6.862 & -- & -- & -- & -- & -- & -- & -- & -- & -- & -- & -- & -- \\
\bottomrule
\end{tabular}}
\caption{Benchmark-balanced means for specialized systems on matched instances. English-source directions average all three benchmarks; reverse directions average FLEURS and NTREX128. toLLMatch uses Llama-3-8B/70B-Instruct; Spanish-to-English extends its original coverage. EAST settings are latency prompts. Bold marks best displayed values, including ties (higher quality, lower LAAL); dashes denote unavailable combinations. Table~\ref{tab:specialized} gives per-benchmark results.}
\label{tab:specialized_mean}
\end{table*}

%% file: sections/conclusion.tex
\section{Conclusion}
\label{sec:conclusion}

SnakeTranslate adapts dual-stream loss-derived alignment to a sampled-history causal lattice and extends it to reverse-KL OPD on student-generated target prefixes. The trained model learns prefix translation and READ/WRITE policies with single-history cached inference. Experiments show improved quality--latency trade-offs over shared-backbone policies and further OPD gains, while larger models retain high-latency quality advantages on WMT24++. Future work targets lattice-training cost, long-input translation, and streaming speech input.

%% file: sections/limitations.tex
\section*{Limitations}

The training lattice is more expensive than a single causal sequence. It contains $O(XY)$ prefix states, and path-restricted attention requires $O(XY(X+Y))$ interactions for fixed model dimensions. Our experiments limit each training stream to 128 positions, so they do not establish the feasibility of lattice training on long documents. Persistent caching removes repeated encoding at inference, but attention still accesses a growing history. The reported LAAL scores measure algorithmic delay, not wall-clock latency or real-time throughput.

Loss-derived alignment provides weak supervision for source sufficiency. Low prediction loss can reflect a language-model prior even when the source has not supplied the relevant information. Consequently, restricting translation loss to the inferred WRITE region does not guarantee faithful translation or eliminate premature predictions. Sampled histories also need not give identical representations of the same source and target prefixes. OPD inherits teacher errors, and the tokenizer and EOS differences described in Appendix~\ref{app:teacher_scoring} introduce an additional mismatch.

Our evaluation uses one 4B backbone and an internal multilingual training corpus. The observed gains need not generalize to other architectures, domains, or language directions. Differences in backbones, data, decoding settings, teacher supervision, and training budgets, together with the absence of an ablation removing offline adaptation, prevent attributing the observed gains to individual components. Generation limits also differ across systems, although capped outputs remain in evaluation. The finite set of operating points and automatic metrics does not establish statistical significance or human-perceived translation quality.

%% file: sections/result_coverage.tex
\section{Additional Results}
\label{app:result_coverage}

\subsection{Direction-Level Results}
\label{sec:fleurs_direction_results}

Table~\ref{tab:target_language} compares seven English-source FLEURS directions on matched instances: Ours at $\tau=0.9$ and external models with LA-2, $c=2$.

\input{tables/target_language_results}

Ours exceeds Riva on both quality metrics in six directions and reduces LAAL in four; Portuguese has slightly lower COMET. Against TranslateGemma on English-to-Chinese, quality rises from 0.715/0.881 to 0.737/0.891 as LAAL rises from 6.281 to 6.668.

On English-to-Japanese, both larger models have higher COMET and lower LAAL, but Ours has higher BLEURT. English-to-Spanish trades quality for delay: Ours reaches 0.749/0.944 at LAAL 3.636 versus Hy-MT2's 0.758/0.955 at 5.557. English-to-Thai is also earlier but weaker than both larger models.

\subsection{Specialized Systems by Benchmark}

Table~\ref{tab:specialized} expands Table~\ref{tab:specialized_mean} by benchmark: English-to-Spanish, Japanese, and Chinese on all three datasets, plus Spanish-to-English and Chinese-to-English on FLEURS/NTREX128. Settings are $k\in\{1,2,3\}$ for toLLMatch/Simul-LLM, native NLLW, $\tau\in\{0.8,0.9,0.95\}$ for Ours, and EAST's \textit{low-medium}, \textit{medium}, and \textit{medium-high} prompts on Chinese--English. Instances match within each direction. Quality coverage is complete; latency alignment has no errors.

\input{tables/specialized_results}

\subsection{Benchmark Analysis}
\label{app:benchmark_analysis}

WMT24++ contains only English-source directions. Matching these directions raises Qwen wait-$3$ LAAL on FLEURS/NTREX128 from 5.368/5.433 to 6.734/6.866, versus 8.110 on WMT24++. Reference-to-source token ratios are 1.329/1.329/1.253 and source lengths approximately 27/26/43. WMT24++ therefore has longer inputs but a smaller length ratio; direction composition alone does not explain its latency. RALCP also depends on variable commitment lengths.

At $\tau=0.95$, Ours trails its offline initialization by 0.002/0.005, 0.007/0.008, and 0.008/0.020 in BLEURT/COMET on the common subsets. The larger WMT24++ COMET gap is consistent with its longer, cross-domain inputs \citep{deutsch2025wmt24pp} and the offline capability gaps in Table~\ref{tab:offline_results}.

\subsection{Direction-Level Stage Gains}
\label{app:stage_details}

OPD improves both quality metrics and LAAL over SFT in all nine dataset--threshold combinations. At $\tau=0.95$, LAAL falls by 10--15\%; BLEURT improves in 39/42, 39/42, and 6/7 directions, and COMET in 40/42, 40/42, and 5/7. English-to-Thai declines on both metrics throughout.

\subsection{Output Statistics}
\label{app:output_statistics}
Across $k\in\{1,2,3\}$, toLLMatch produces 39,030 predictions with 12 capped, none empty, and complete source consumption. Simul-LLM produces 10,800 with none capped or empty and complete source consumption. LA produces 1,002,093 predictions across supported directions and three read increments: 14 empty and 497 with incomplete source consumption. EAST produces 32,725 predictions across five prompts, none empty. Its 38 per-WRITE budget hits resume READ and do not truncate the whole translation. All nonempty streaming predictions have valid latency alignment.

Offline Base evaluation has 204/121,266 capped outputs (0.168\%) and none empty. Offline-tuned Qwen has 20 capped (0.016\%) and 267 empty (0.220\%) on the same instances. Riva, TranslateGemma, and Hy-MT2 have 5/44,466, 1/47,033, and 1/121,266 capped outputs, respectively, and none empty. These counts cover each model's full supported set, not the offline table's common subset. Empty and capped outputs retain quality scores; empty outputs have undefined latency.

%% file: tables/target_language_results.tex
\begin{table*}[tp]
\centering\scriptsize
\setlength{\tabcolsep}{3pt}
\renewcommand{\arraystretch}{1.05}
\begin{tabular}{lllccccccc}
\toprule
System & Policy & Metric & en$\to$zh & en$\to$ja & en$\to$ko & en$\to$th & en$\to$es & en$\to$pt & en$\to$id \\
\midrule
Ours & $\tau=0.9$ & BLEURT & 0.737 & 0.729 & 0.719 & 0.688 & 0.749 & 0.781 & 0.816 \\
 &  & COMET & 0.891 & 0.912 & 0.908 & 0.875 & 0.944 & 0.944 & 0.932 \\
 &  & LAAL & 6.668 & 7.718 & 8.320 & 4.557 & 3.636 & 3.071 & 3.340 \\
\midrule
Riva-4B & Local Agreement ($c=2$) & BLEURT & 0.708 & 0.703 & 0.679 & 0.619 & 0.743 & 0.775 & 0.794 \\
 &  & COMET & 0.852 & 0.891 & 0.874 & 0.785 & 0.939 & 0.947 & 0.913 \\
 &  & LAAL & 6.471 & 7.385 & 7.729 & 6.432 & 5.335 & 4.939 & 5.271 \\
\midrule
TranslateGemma-27B & Local Agreement ($c=2$) & BLEURT & 0.715 & 0.711 & 0.715 & 0.702 & 0.743 & 0.767 & 0.815 \\
 &  & COMET & 0.881 & 0.920 & 0.909 & 0.914 & 0.953 & 0.964 & 0.956 \\
 &  & LAAL & 6.281 & 6.917 & 7.288 & 5.352 & 5.254 & 5.117 & 5.190 \\
\midrule
Hy-MT2-30B-A3B & Local Agreement ($c=2$) & BLEURT & 0.736 & 0.711 & 0.702 & 0.703 & 0.758 & 0.784 & 0.819 \\
 &  & COMET & 0.904 & 0.922 & 0.910 & 0.915 & 0.955 & 0.958 & 0.953 \\
 &  & LAAL & 6.713 & 7.443 & 7.598 & 5.574 & 5.557 & 5.073 & 5.428 \\
\bottomrule
\end{tabular}
\caption{FLEURS results on seven common English-source directions. Ours uses $\tau=0.9$; external models use Local Agreement with $c=2$. Columns report individual directions on identical evaluation instances. Higher BLEURT/COMET and lower LAAL are preferable.}
\label{tab:target_language}
\end{table*}

%% file: tables/specialized_results.tex
\begin{table*}[tp]
\centering\scriptsize
\setlength{\tabcolsep}{1.5pt}
\resizebox{\textwidth}{!}{%
\begin{tabular}{llccccccccccccccc}
\toprule
System & Setting & \multicolumn{3}{c}{en$\to$es} & \multicolumn{3}{c}{en$\to$ja} & \multicolumn{3}{c}{en$\to$zh} & \multicolumn{3}{c}{es$\to$en} & \multicolumn{3}{c}{zh$\to$en} \\
\cmidrule(lr){3-5}\cmidrule(lr){6-8}\cmidrule(lr){9-11}\cmidrule(lr){12-14}\cmidrule(lr){15-17}
 & & BLEURT & XCOMET & LAAL & BLEURT & XCOMET & LAAL & BLEURT & XCOMET & LAAL & BLEURT & XCOMET & LAAL & BLEURT & XCOMET & LAAL \\
\midrule
\addlinespace[8pt]
\multicolumn{17}{c}{\textbf{FLEURS}}\\
\midrule
Ours & $\tau=0.8$ & 0.737 & 0.930 & \textbf{2.938} & 0.722 & 0.900 & \textbf{6.441} & 0.727 & 0.875 & 5.556 & 0.752 & 0.947 & \textbf{3.020} & 0.759 & 0.946 & 4.348 \\
 & $\tau=0.9$ & 0.749 & 0.944 & 3.636 & 0.729 & 0.912 & 7.718 & 0.737 & \textbf{0.891} & 6.668 & 0.760 & 0.957 & 4.026 & 0.764 & 0.952 & 5.526 \\
 & $\tau=0.95$ & \textbf{0.753} & \textbf{0.947} & 4.324 & \textbf{0.731} & \textbf{0.916} & 9.192 & \textbf{0.741} & \textbf{0.891} & 7.861 & \textbf{0.764} & \textbf{0.959} & 4.930 & \textbf{0.769} & \textbf{0.956} & 6.696 \\
\midrule
toLLMatch-70B & $k=1$ & 0.719 & 0.898 & 5.366 & -- & -- & -- & -- & -- & -- & 0.735 & 0.928 & 3.976 & -- & -- & -- \\
 & $k=2$ & 0.732 & 0.920 & 6.251 & -- & -- & -- & -- & -- & -- & 0.740 & 0.934 & 4.627 & -- & -- & -- \\
 & $k=3$ & 0.737 & 0.929 & 7.350 & -- & -- & -- & -- & -- & -- & 0.747 & 0.944 & 5.526 & -- & -- & -- \\
\midrule
toLLMatch-8B & $k=1$ & 0.682 & 0.853 & 5.367 & -- & -- & -- & -- & -- & -- & 0.679 & 0.833 & 3.639 & -- & -- & -- \\
 & $k=2$ & 0.706 & 0.889 & 6.280 & -- & -- & -- & -- & -- & -- & 0.714 & 0.885 & 4.472 & -- & -- & -- \\
 & $k=3$ & 0.716 & 0.907 & 7.390 & -- & -- & -- & -- & -- & -- & 0.733 & 0.912 & 5.428 & -- & -- & -- \\
\midrule
EAST-8B & low-medium & -- & -- & -- & -- & -- & -- & 0.674 & 0.749 & \textbf{5.198} & -- & -- & -- & 0.704 & 0.865 & \textbf{3.252} \\
 & medium & -- & -- & -- & -- & -- & -- & 0.691 & 0.784 & 7.306 & -- & -- & -- & 0.727 & 0.900 & 4.633 \\
 & medium-high & -- & -- & -- & -- & -- & -- & 0.701 & 0.808 & 10.240 & -- & -- & -- & 0.736 & 0.918 & 5.727 \\
\midrule
NLLW & -- & 0.601 & 0.704 & 6.467 & 0.540 & 0.654 & 10.714 & 0.558 & 0.672 & 8.880 & -- & -- & -- & 0.681 & 0.827 & 19.254 \\
\midrule
Simul-LLM & $k=1$ & 0.526 & 0.606 & 4.523 & -- & -- & -- & -- & -- & -- & -- & -- & -- & -- & -- & -- \\
 & $k=2$ & 0.570 & 0.687 & 5.310 & -- & -- & -- & -- & -- & -- & -- & -- & -- & -- & -- & -- \\
 & $k=3$ & 0.595 & 0.731 & 6.419 & -- & -- & -- & -- & -- & -- & -- & -- & -- & -- & -- & -- \\
\midrule
\addlinespace[8pt]
\multicolumn{17}{c}{\textbf{NTREX128}}\\
\midrule
Ours & $\tau=0.8$ & 0.740 & 0.895 & \textbf{3.002} & 0.656 & 0.839 & \textbf{6.759} & 0.672 & 0.807 & 5.555 & 0.759 & 0.934 & \textbf{3.016} & 0.706 & 0.889 & 4.930 \\
 & $\tau=0.9$ & 0.755 & 0.914 & 3.697 & 0.666 & 0.853 & 8.182 & 0.680 & 0.822 & 6.847 & 0.772 & 0.946 & 3.996 & 0.714 & 0.902 & 6.207 \\
 & $\tau=0.95$ & \textbf{0.761} & \textbf{0.921} & 4.482 & \textbf{0.669} & \textbf{0.860} & 9.754 & \textbf{0.684} & \textbf{0.829} & 8.277 & \textbf{0.778} & \textbf{0.952} & 4.972 & \textbf{0.716} & \textbf{0.905} & 7.436 \\
\midrule
toLLMatch-70B & $k=1$ & 0.717 & 0.869 & 5.199 & -- & -- & -- & -- & -- & -- & 0.743 & 0.918 & 3.730 & -- & -- & -- \\
 & $k=2$ & 0.738 & 0.894 & 6.033 & -- & -- & -- & -- & -- & -- & 0.752 & 0.928 & 4.419 & -- & -- & -- \\
 & $k=3$ & 0.745 & 0.907 & 7.148 & -- & -- & -- & -- & -- & -- & 0.760 & 0.934 & 5.348 & -- & -- & -- \\
\midrule
toLLMatch-8B & $k=1$ & 0.659 & 0.785 & 5.200 & -- & -- & -- & -- & -- & -- & 0.695 & 0.835 & 3.711 & -- & -- & -- \\
 & $k=2$ & 0.692 & 0.834 & 6.121 & -- & -- & -- & -- & -- & -- & 0.727 & 0.881 & 4.489 & -- & -- & -- \\
 & $k=3$ & 0.711 & 0.862 & 7.218 & -- & -- & -- & -- & -- & -- & 0.742 & 0.904 & 5.397 & -- & -- & -- \\
\midrule
EAST-8B & low-medium & -- & -- & -- & -- & -- & -- & 0.621 & 0.695 & \textbf{5.393} & -- & -- & -- & 0.654 & 0.791 & \textbf{3.376} \\
 & medium & -- & -- & -- & -- & -- & -- & 0.636 & 0.731 & 7.546 & -- & -- & -- & 0.678 & 0.837 & 4.739 \\
 & medium-high & -- & -- & -- & -- & -- & -- & 0.646 & 0.752 & 11.249 & -- & -- & -- & 0.687 & 0.852 & 5.950 \\
\midrule
NLLW & -- & 0.605 & 0.660 & 6.637 & 0.461 & 0.577 & 10.893 & 0.483 & 0.611 & 9.187 & -- & -- & -- & 0.637 & 0.762 & 19.998 \\
\midrule
Simul-LLM & $k=1$ & 0.515 & 0.568 & 4.429 & -- & -- & -- & -- & -- & -- & -- & -- & -- & -- & -- & -- \\
 & $k=2$ & 0.563 & 0.643 & 5.353 & -- & -- & -- & -- & -- & -- & -- & -- & -- & -- & -- & -- \\
 & $k=3$ & 0.587 & 0.678 & 6.432 & -- & -- & -- & -- & -- & -- & -- & -- & -- & -- & -- & -- \\
\midrule
\addlinespace[8pt]
\multicolumn{17}{c}{\textbf{WMT24++}}\\
\midrule
Ours & $\tau=0.8$ & 0.651 & 0.801 & \textbf{3.359} & 0.603 & 0.764 & \textbf{6.155} & 0.643 & 0.756 & \textbf{5.322} & -- & -- & -- & -- & -- & -- \\
 & $\tau=0.9$ & 0.670 & 0.821 & 4.120 & 0.612 & 0.780 & 7.348 & 0.647 & 0.772 & 6.464 & -- & -- & -- & -- & -- & -- \\
 & $\tau=0.95$ & \textbf{0.678} & \textbf{0.828} & 5.058 & \textbf{0.616} & \textbf{0.787} & 8.875 & \textbf{0.657} & \textbf{0.781} & 7.894 & -- & -- & -- & -- & -- & -- \\
\midrule
toLLMatch-70B & $k=1$ & 0.597 & 0.748 & 7.052 & -- & -- & -- & -- & -- & -- & -- & -- & -- & -- & -- & -- \\
 & $k=2$ & 0.606 & 0.763 & 7.876 & -- & -- & -- & -- & -- & -- & -- & -- & -- & -- & -- & -- \\
 & $k=3$ & 0.612 & 0.774 & 8.924 & -- & -- & -- & -- & -- & -- & -- & -- & -- & -- & -- & -- \\
\midrule
toLLMatch-8B & $k=1$ & 0.549 & 0.680 & 7.101 & -- & -- & -- & -- & -- & -- & -- & -- & -- & -- & -- & -- \\
 & $k=2$ & 0.566 & 0.707 & 7.989 & -- & -- & -- & -- & -- & -- & -- & -- & -- & -- & -- & -- \\
 & $k=3$ & 0.574 & 0.720 & 8.990 & -- & -- & -- & -- & -- & -- & -- & -- & -- & -- & -- & -- \\
\midrule
EAST-8B & low-medium & -- & -- & -- & -- & -- & -- & 0.573 & 0.645 & 5.899 & -- & -- & -- & -- & -- & -- \\
 & medium & -- & -- & -- & -- & -- & -- & 0.588 & 0.678 & 7.788 & -- & -- & -- & -- & -- & -- \\
 & medium-high & -- & -- & -- & -- & -- & -- & 0.594 & 0.693 & 9.344 & -- & -- & -- & -- & -- & -- \\
\midrule
NLLW & -- & 0.489 & 0.568 & 7.252 & 0.347 & 0.501 & 12.618 & 0.429 & 0.545 & 12.178 & -- & -- & -- & -- & -- & -- \\
\midrule
Simul-LLM & $k=1$ & 0.422 & 0.472 & 5.582 & -- & -- & -- & -- & -- & -- & -- & -- & -- & -- & -- & -- \\
 & $k=2$ & 0.456 & 0.510 & 6.637 & -- & -- & -- & -- & -- & -- & -- & -- & -- & -- & -- & -- \\
 & $k=3$ & 0.469 & 0.529 & 7.734 & -- & -- & -- & -- & -- & -- & -- & -- & -- & -- & -- & -- \\
\bottomrule
\end{tabular}}
\caption{Per-direction results for specialized simultaneous-translation systems, grouped by benchmark. Columns cover the union of their evaluated directions and report BLEURT, XCOMET, and LAAL. Within each benchmark, bold indicates column-best displayed values (higher quality, lower LAAL), including ties. Dashes indicate unavailable combinations. NLLW has no $k$ parameter; EAST settings denote its native latency prompts.}
\label{tab:specialized}
\end{table*}

%% file: sections/appendix_cases.tex
\section{Case Studies}
\label{app:case_studies}

\subsection{Loss-Derived Alignment}
\label{sec:alignment_case}

\begin{figure*}[p]
 \centering
 \includegraphics[width=0.95\textwidth]{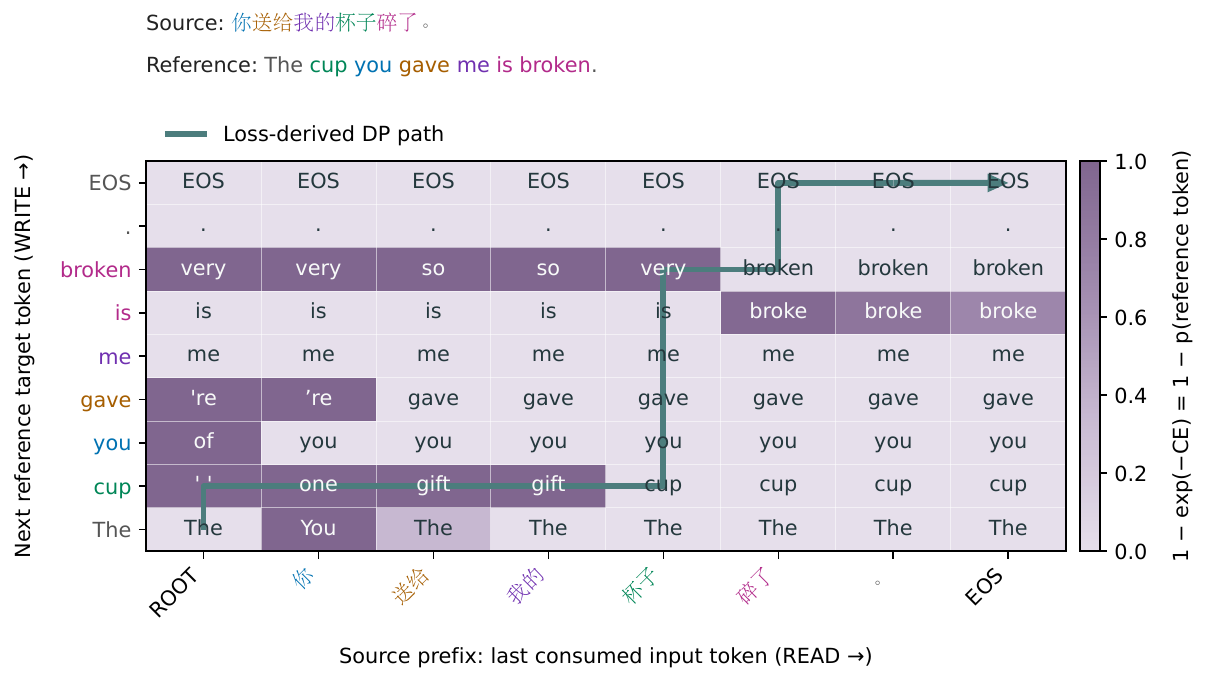}
 \caption{Teacher-forced Chinese-to-English alignment diagnostic, not a free-decoding trajectory. Matching text colors indicate corresponding expressions. Columns index source prefixes; rows index next reference tokens. Cells show top predictions; darker purple means larger $1-\exp(-c_{x,y}^{\mathrm{SFT}})$. Axes retain tokenizer subwords. The teal DP path joins cell centers, projecting terminal states onto the final row or column.}
 \label{fig:alignment_case}
\end{figure*}

Figure~\ref{fig:alignment_case} uses the constructed sentence \textit{The cup you gave me is broken.} Chinese places \textit{you gave me} before \textit{cup}; English places the relative clause after it. The grid comes from a teacher-forced forward pass of Ours using the training computation, without parameter updates.

The path reads \textit{cup} before writing \textit{cup you gave me is}, then reads the predicate before writing \textit{broken}. Their reference-token cross-entropies fall from 24.630 to $2.94\times10^{-4}$ and 16.500 to $4.01\times10^{-4}$ across the relevant source tokens. Every top prediction at a path WRITE cell matches the reference. This teacher-forced diagnostic illustrates prediction difficulty as a weak timing signal, but its alignment remains approximate: articles, the copula, and EOS can precede their source counterparts.

\subsection{Streaming Decisions}
\label{app:streaming_case}

\begin{figure*}[p]
 \centering
 \includegraphics[width=\textwidth]{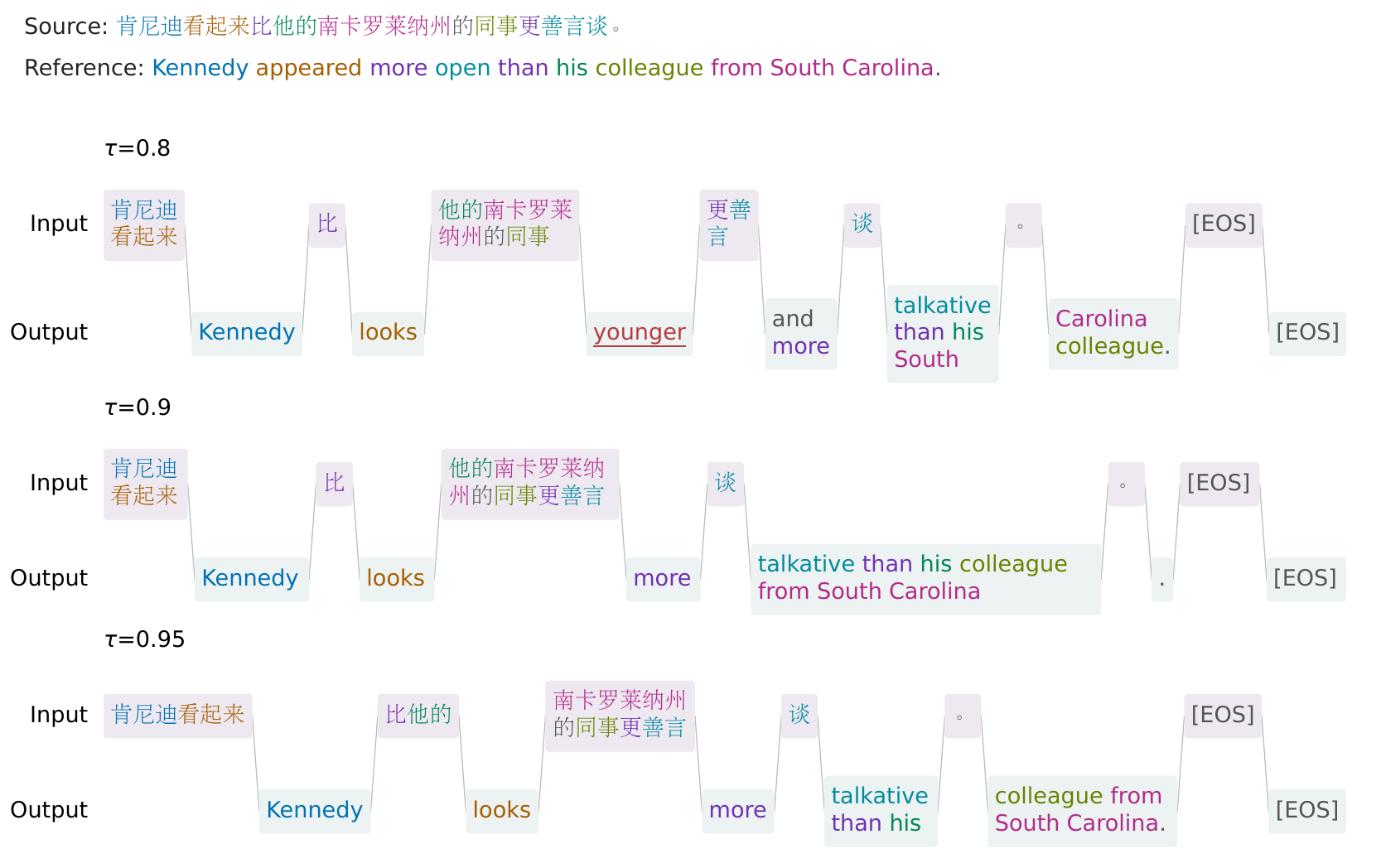}
 \caption{Streaming translation at three thresholds on a Chinese-to-English NTREX128 example. Matching colors indicate corresponding expressions; red underlined text marks an unsupported prediction. Input and output segments are interleaved in READ/WRITE order.}
 \label{fig:threshold_case}
\end{figure*}

Figure~\ref{fig:threshold_case} shows free decoding on a Chinese-to-English NTREX128 example. The source places \textit{more talkative} after the phrase identifying Kennedy's colleague. At $\tau=0.8$, Ours writes \textit{younger} at source position eleven, before reading the property, followed by \textit{and more talkative}. At $\tau=0.9$, it reads to position fourteen, where the property begins, and writes \textit{more talkative} without the unsupported age comparison. At $\tau=0.95$, the translation is identical, but \textit{looks} and \textit{colleague from South Carolina} each arrive one source subword later. Positions count model source subwords, including EOS.

%% file: sections/appendix_architecture.tex
\section{Implementation and Evaluation}
\label{app:architecture}

\subsection{Model and Alignment}
\label{app:execution_cost}

The READ/WRITE gate comprises layer normalization and a two-layer MLP with a 256-dimensional bottleneck, a sigmoid linear unit (SiLU) activation, dropout 0.1, and two action logits.

For a padded batch of size $B$ and hidden width $D$, the inter-layer representation has shape $B\times(X+1)\times Y\times D$ under the prompt-suppressed indexing of Section~\ref{sec:model_architecture}. Each cell carries one hidden vector; recurrent states and attention keys and values are local to each layer.

For fixed model dimensions, recurrence requires $O(XY)$ work per example, with equal-depth nodes processed in parallel; attention requires $O(XY(X+Y))$ query--ancestor interactions. Projections and feed-forward operations batch across cells. Attention-query and vocabulary-loss chunking bounds temporary memory without removing cells from the training lattice. These costs omit hidden-width and vocabulary-size factors.

Incremental inference retains recurrent states, convolution histories, and attention keys and values without recomputing historical hidden states. For history length $L$, recurrent-state storage is constant, KV storage is $O(L)$, and attention requires $O(L)$ work per step and $O(L^2)$ over the sequence.

\subsection{Relation to Prior Work}

Table~\ref{tab:prior_comparison} distinguishes the inherited alignment objective from the sampled-history computation and distillation extension. Both models use prediction costs to recover a monotone boundary and train the policy throughout the lattice. SnakeTranslate changes the representation and execution associated with each cell and extends the cost to teacher distributions on student-generated target prefixes.

\begin{table*}[tp]
 \centering\small
 \setlength{\tabcolsep}{4pt}
 \begin{tabular}{@{}p{.17\textwidth}p{.35\textwidth}p{.43\textwidth}@{}}
 \toprule
 Component & Dual-stream grid \citep{pu2026dualstream} & SnakeTranslate\\
 \midrule
 Cell representation & Separate source- and target-stream hidden states & One hidden state for a sampled interleaved history\\
 Layer computation & Four self-/cross-stream interactions with broadcast and cellwise-score approximations & Causal recurrence and attention along the sampled ancestor path\\
 Boundary objective & Reference-token CE balanced against an early-output area reward & Inherited objective; reverse KL replaces CE during OPD\\
 Supervision region & Policy loss throughout the grid; translation loss below the boundary & Inherited region-based supervision on sampled causal histories\\
 Cached inference & Extend a source row or target column of the dual-stream frontier & Append one token to a single history and update ordinary recurrent/KV states\\
 Distillation & No student-rollout reverse-KL extension & Full-source teacher distributions supervise both boundary inference and translation on student-generated target prefixes\\
 \bottomrule
 \end{tabular}
 \caption{Relationship to the dual-stream grid framework. Grid-based alignment and region supervision are inherited; sampled-history causal execution and its reverse-KL extension are the focus of this work.}
 \label{tab:prior_comparison}
\end{table*}

%% file: sections/appendix_alignment.tex
\label{app:alignment_details}

Equation~\ref{eq:dp} initializes $F(0,0)=0$ and disallows out-of-range predecessors. Backtracking favors WRITE on ties.

DP closes with a sentinel row that repeats the final cost row without consuming source tokens. Moving a sentinel-row WRITE before the terminal READ preserves prediction cost and increases area reward for $\lambda>0$, so all targets are optimally written by row $X$. Adding the constant $\lambda Y$ to the terminal score recovers Eq.~\ref{eq:path_objective}.

The predecessor sampler counts target-prompt positions, whereas DP indexes the target suffix; the main notation suppresses this offset. A shared root initializes the history. Both training and cached inference assign position zero to each stream's first token.

SFT pools valid cells across the batch, with a separate DP boundary per example. The translation term weights examples by their selected-cell counts, while the gate term weights them by their valid-cell counts. OPD normalizes each term within each example before averaging, giving examples equal weight. Padding and target-prompt positions contribute to neither loss; an empty translation selection contributes zero. Path costs use WRITE edges, while the resulting boundary supervises a full region. Backtracking is detached; READ cells can still receive gradients through the gate and shared computations.

DP sums the unaveraged per-token CE or full-vocabulary KL costs along WRITE edges. Loss averaging occurs only after boundary recovery. The area weight $\lambda$ is held fixed within each search and measured relative to the corresponding cost scale; equal numerical values need not induce equal delays under CE and KL.

\subsection{Training Procedure}

Table~\ref{tab:training_procedure} summarizes the shared computation. A rollout supplies target content; the training tree is sampled anew rather than copied from the rollout's READ/WRITE trajectory.

\begin{table*}[tp]
 \centering\small
 \begin{tabular}{@{}rp{.92\textwidth}@{}}
 \toprule
 Step & One training update\\
 \midrule
 1 & Sample source $s$ and set target $z=t$ for SFT, or generate $z=\operatorname{Rollout}_{\theta_k}(s)$ for OPD.\\
 2 & Sample one valid parent per lattice cell, shared across layers. Embed the corresponding source or target token, including root and prompt context.\\
 3 & Compute each $h_{x,y}$ by ancestor-restricted causal layers, then obtain $p_{x,y}$ and $a_{x,y}$. In OPD, obtain fixed full-source teacher distributions $q_y$ on $z$ in one causal pass.\\
 4 & Form $c_{x,y}=-\log p_{x,y}(z_{y+1})$ for SFT, or $D_{\mathrm{KL}}(p_{x,y}\Vert q_y)$ for OPD. Use $\operatorname{stopgrad}(c)$ in DP to recover $\pi^\star$.\\
 5 & Derive $g_x^\star$ and $M_{x,y}=\mathbf1[y<g_x^\star]$. Keep the recovered mask fixed during backpropagation.\\
 6 & Sum translation loss on WRITE-region cells and gate loss on all valid cells, with the stage-specific averaging described above. Backpropagate through student costs, gate predictions, and shared ancestor computations.\\
 7 & Update student parameters. In OPD, synchronize rollout weights after each optimizer update, allowing at most one update of asynchronous lag.\\
 \bottomrule
 \end{tabular}
 \caption{SFT and OPD training procedure. Parent sampling, teacher scoring, rollout actions, and DP backtracking receive no gradients. Chunking changes execution order, not the set of supervised cells.}
 \label{tab:training_procedure}
\end{table*}

%% file: sections/implementation_details.tex
\subsection{Training and Distillation}
\label{app:dp_details}
\label{app:opd_reduction}
\label{app:task_formatting}
\label{app:teacher_scoring}

SFT and OPD use bfloat16 (BF16) computation and DeepSpeed's Zero Redundancy Optimizer (ZeRO) stage 2 \citep{rajbhandari2020zero}. SFT uses activation checkpointing and chunking on complete grids. OPD uses AdamW with $\beta=(0.9,0.999)$, $\epsilon=10^{-8}$, weight decay 0.01, global gradient-norm clipping at 1.0, and single-precision (FP32) gradient accumulation. Rollout weights synchronize after every optimizer update, with at most one update of asynchronous lag. Learning rates, effective batch sizes, and stage initializations are given in Section~\ref{sec:training_setup}.

OPD applies Eqs.~\ref{eq:gate_loss} and~\ref{eq:translation_loss} separately to each example, then averages over batch size $B$:
\begin{equation}
 \mathcal L_{\mathrm{OPD}}=\frac{1}{B}\sum_{b=1}^{B}
 \left(\mathcal L_{\mathrm{trans}}^{(b)}
       +\mathcal L_{\mathrm{gate}}^{(b)}\right).
\end{equation}
Both terms have unit weight. Student logits are converted to FP32 before log-softmax. Reverse KL sums over the full vocabulary at temperature 1, without top-$k$/top-$p$ truncation or clipping, then averages over each example's DP-selected WRITE cells. Teacher probabilities are fixed; student gradients flow through both the weighting and log-ratio in Eq.~\ref{eq:reverse_kl}. No reference-token loss or student-continuation hard-label loss is added. Samples exceeding either 128-position stream limit are discarded, not truncated.

Source text is followed by EOS. The target stream is \texttt{Translate \{source language\} to \{target language\}:}, directly followed by the translation and EOS, without a newline or other separator. Prompt tokens provide context without translation or policy loss.

The frozen internal Qwen3.8 teacher receives the complete source in its native chat template with reasoning disabled. Its user message is:
\begin{quote}\small
Translate \{src\} to \{tgt\} for simultaneous interpretation.

Keep the translation faithful, natural, and idiomatic in \{tgt\}, preserving the source's tone and register. Avoid large-scale reordering of information where possible, while allowing local reordering for natural target-language expression.

Your final answer should contain only the translation.

Source:\\
\{source\}
\end{quote}
Here \{src\} and \{tgt\} are language names. Student token IDs are appended directly to the assistant prefix without re-tokenization. One teacher-forced pass supplies next-token distributions without reference translations. The sampled training tree is not constrained to follow the rollout's execution history.

Student and teacher share base-vocabulary token IDs but differ in pre-tokenization and some special tokens. Scoring retains the student's EOS despite the teacher's different native EOS, leaving a termination-semantic mismatch.

%% file: sections/appendix_evaluation.tex
\subsection{Evaluation Protocol}
\label{app:evaluation_details}
\label{app:baseline_protocols}
\label{app:latency_protocol}

The 42 directions connect Chinese, English, and Japanese mutually and bidirectionally with Spanish, Indonesian, Korean, Malay, Portuguese, and Thai. FLEURS pairs test transcriptions by sentence ID, excludes unmatched entries, and retains repeated transcriptions from distinct recordings, yielding 30,672 instances. NTREX128 supplies 1,997 aligned items per direction (83,874 total); non-English sources are human translations of English originals. WMT24++ covers the seven English-source directions in Table~\ref{tab:target_language}. Excluding sources marked as bad leaves 960 items per direction (6,720 total), scored against released references. Evaluation sources are never truncated.

Table~\ref{tab:coverage} summarizes system coverage. Shared-backbone comparisons use 42/42/7 directions; external-model and offline comparisons use the common 13/13/7. On FLEURS and NTREX128, this intersection comprises both directions between English and Chinese, Japanese, Korean, Thai, Spanish, and Indonesian, plus English-to-Portuguese. WMT24++ uses all seven English-source directions.

\begin{table}[tp]
 \centering\small
 \setlength{\tabcolsep}{3pt}
 \begin{tabular}{lrrr}
 \toprule
 System & FLEURS & NTREX & WMT\\
 \midrule
 \input{tables/coverage_rows}
 \end{tabular}
 \caption{Evaluated directions per system and benchmark. The four offline backbones also support wait-$k$, RALCP, and LA. Main comparisons use matched subsets.}
 \label{tab:coverage}
\end{table}

Wait-$k$ and RALCP use $k\in\{3,6,9\}$; RALCP uses five candidates and acceptance ratio 0.6. LA uses beam size 2 (LA-2) with read increments $c\in\{1,2,3\}$ in each backbone's native source tokens. It commits the common uncommitted prefix of consecutive best hypotheses and flushes the remaining hypothesis after source completion.

EAST-8B uses released fine-tuned weights, its native interleaved policy, WRITE beam size five, and native latency prompts. Table~\ref{tab:specialized} reports \textit{low-medium}, \textit{medium}, and \textit{medium-high}. READ units are whitespace-delimited English words or tokenizer-derived valid Unicode fragments for Chinese. Each WRITE segment has a budget of $\min(1024,2(r+25))$ tokens, where $r$ counts source units read since the preceding WRITE segment and resets after each WRITE segment. Budget exhaustion resumes READ without terminating translation.

toLLMatch uses Meta-Llama-3-8B-Instruct and Meta-Llama-3-70B-Instruct, their respective tokenizers, and the authors' prompt template, system message, and response priming. Decoding is greedy, with an 8,192-token context. We preserve sentence-final punctuation stopping and the original vLLM stopping-string boundary behavior, including minimum generation lengths. The source clock advances one word after every policy call, including WRITE. English-to-Spanish is the original direction; Spanish-to-English extends evaluation on FLEURS and NTREX128.

Both toLLMatch variants and Simul-LLM use native $k\in\{1,2,3\}$ settings. In toLLMatch, generation is first attempted after reading more than $k$ source words unless input is complete. Simul-LLM instead requires an ongoing source--target word-count difference of $k$ for WRITE. Their traces are mapped to common subword units; equal $k$ values do not imply equal schedules.

The five offline controls in Table~\ref{tab:offline_results} use complete sources, greedy vLLM decoding, and model-specific end-of-turn stopping on the common 13/13/7 directions. Qwen models use the offline-training prompt; external models retain their own prompts. Output caps are 2,048 tokens for controls and most streaming baselines, versus 256/512 for SnakeTranslate on FLEURS/other benchmarks, subject to native context limits and EAST's per-WRITE budgets.

From saved traces, WRITE positions are mapped to common FLoRes SentencePiece units using the released \texttt{flores200sacrebleuspm} model.
\newpage
We tokenize each complete source, hypothesis, and reference, then recover the UTF-8 byte boundary of each common subword. Each target subword inherits the read position of the first WRITE event whose committed bytes cover its right boundary. An event may complete several subwords; incomplete Unicode characters spanning native tokens are not counted early. Source consumption counts only common subwords fully covered by the reconstructed native prefix. EOS and other special tokens add no text units.

Let $g(t)$ be the source length consumed when writing target subword $t$, and $S$, $H$, and $R$ the source, hypothesis, and reference lengths. Then
\begin{equation}
 \operatorname{LAAL}=\frac{1}{T}\sum_{t=1}^{T}
 \left[g(t)-\frac{(t-1)S}{\max(H,R)}\right],
\end{equation}
where $T$ is the first target position after complete source consumption, or $H$ if none exists. A terminal EOS adds no target unit: if no text WRITE observes source completion, all $H$ units are used. Negative LAAL is retained; incomplete source consumption is flagged and retained. Empty outputs have undefined latency and are counted separately. Valid latencies are averaged within directions, then equally across directions. LAAL excludes computation time.

BLEURT-20 and XCOMET-XL use the same direction-balanced averaging, retaining empty and truncated outputs. Appendix~\ref{app:output_statistics} reports output statistics.

%% file: tables/coverage_rows.tex
Ours: OPD & 42 & 42 & 7 \\
Ours: SFT & 42 & 42 & 7 \\
EAST-8B & 2 & 2 & 1 \\
Hy-MT2-30B-A3B & 42 & 42 & 7 \\
NLLW & 4 & 4 & 3 \\
Offline-tuned Qwen3.5-4B & 42 & 42 & 7 \\
Qwen3.5-4B-Base & 42 & 42 & 7 \\
Riva-Translate-4B & 14 & 14 & 7 \\
Simul-LLM & 1 & 1 & 1 \\
TranslateGemma-27B & 15 & 15 & 7 \\
toLLMatch-70B & 2 & 2 & 1 \\
toLLMatch-8B & 2 & 2 & 1 \\
\bottomrule